\documentclass[sigconf,natbib=true,anonymous=false]{acmart}
\usepackage{multirow}
\usepackage{algorithm}
\usepackage{algorithmic}
\usepackage{booktabs}
\usepackage{xcolor}

\newcommand{\ie}{\textit{i.e.}}
\newcommand{\eg}{\textit{e.g.}}
\newcommand{\ttt}{\texttt}

\AtBeginDocument{%
  \providecommand\BibTeX{{%
    \normalfont B\kern-0.5em{\scshape i\kern-0.25em b}\kern-0.8em\TeX}}}

\setcopyright{acmcopyright}
\copyrightyear{2018}
\acmYear{2018}
\acmDOI{XXXXXXX.XXXXXXX}

\acmPrice{15.00}
\acmISBN{978-1-4503-XXXX-X/18/06}

\begin{document}

\title{PReGAN: Answer Oriented Passage Ranking with Weakly Supervised GAN}


\author{Pan Du}
\email{du@youark.com}
\authornotemark[1]
\affiliation{%
  \institution{Thomson Reuters Labs}
  \city{Toronto}
  \country{Canada}
}

\author{Jian-Yun Nie}
\affiliation{%
  \institution{University of Montreal}
  \city{Montreal}
  \country{Canada}}
\email{nie@iro.umontreal.ca}

\author{Yutao Zhu}
\affiliation{%
  \institution{University of Montreal}
  \city{Montreal}
  \country{Canada}}
\email{yutaozhu94@gmail.com}

\author{Hao Jiang}
\affiliation{%
  \institution{Huawei Poisson Lab.}
  \city{Shenzhen}
  \country{China}}
\email{jianghao66@huawei.com}

\author{Lixin Zou}
\affiliation{%
  \institution{Tshinghua Univeristy}
  \city{Beijing}
  \country{China}}
\email{zoulx15@mails.tsinghua.edu.cn}

\author{Xiaohui Yan}
\affiliation{%
  \institution{Huawei Poisson Lab.}
  \city{Shenzhen}
  \country{China}}
\email{yanxiaohui2@huawei.com}



\begin{abstract}
Beyond topical relevance, passage ranking for open-domain factoid question answering also requires a passage to contain an answer (answerability). While a few recent studies have incorporated some reading capability into a ranker to account for answerability, the ranker is still hindered by the noisy nature of the training data typically available in this area, which considers any passage containing an answer entity as a positive sample. However, the answer entity in a passage is not necessarily mentioned in relation with the given question. To address the problem, we propose an approach called \ttt{PReGAN} for Passage Reranking based on Generative Adversarial Neural networks, which incorporates a discriminator on answerability, in addition to a  discriminator  on topical relevance. The goal is to  force the generator to rank  higher a passage that is topically relevant and  contains an answer. Experiments on five public datasets show that \ttt{PReGAN} can better rank appropriate passages, which in turn, boosts the effectiveness of QA systems, and outperforms the existing approaches without using external data.
\end{abstract}





\maketitle

\section{Introduction}
A common type of open-domain question answering (OpenQA) aims to find answers from a large collection of texts~\cite{chen2017reading}. It operates generally in two steps: finding a limited number of candidate passages using a retrieval method, and performing machine reading through these texts to extract answers. 
The state-of-the-art machine reading methods have produced human-level performance~\cite{zhang2020retrospective} when reading the \textit{right} passage. However, when reading the passages \textit{retrieved} with the question, the performance drops dramatically~\cite{min2021neurips,karpukhin2020dense}, showing the critical importance of retrieving good candidate passages. Many studies have been devoted to improving the topical relevance of retrieved candidate passages~\cite{lin2018denoising, karpukhin2020dense, wang2018r, lee2019latent}, but few studies have investigated the problem of \textbf{answerability}, \ie, whether a retrieved passage may contain an answer. Both criteria - topical relevance and answerability, are critical in the context of OpenQA. In fact, a passage highly relevant to the question may not necessarily contain an answer, and a passage that contains an answer entity may not be relevant to the question. In both cases, the reader may be misled by these passages to select a wrong answer, as the reader highly relies on the few passages provided to it. It is thus important that the selected passages for the reader should both be relevant and contain an asnwer.

In this paper, we propose an approach to refine the list of candidate passages according to both criteria 
through an answer-oriented passage (re-)ranking.

\begin{figure}[t]
    \centering
    \includegraphics[width=\linewidth]{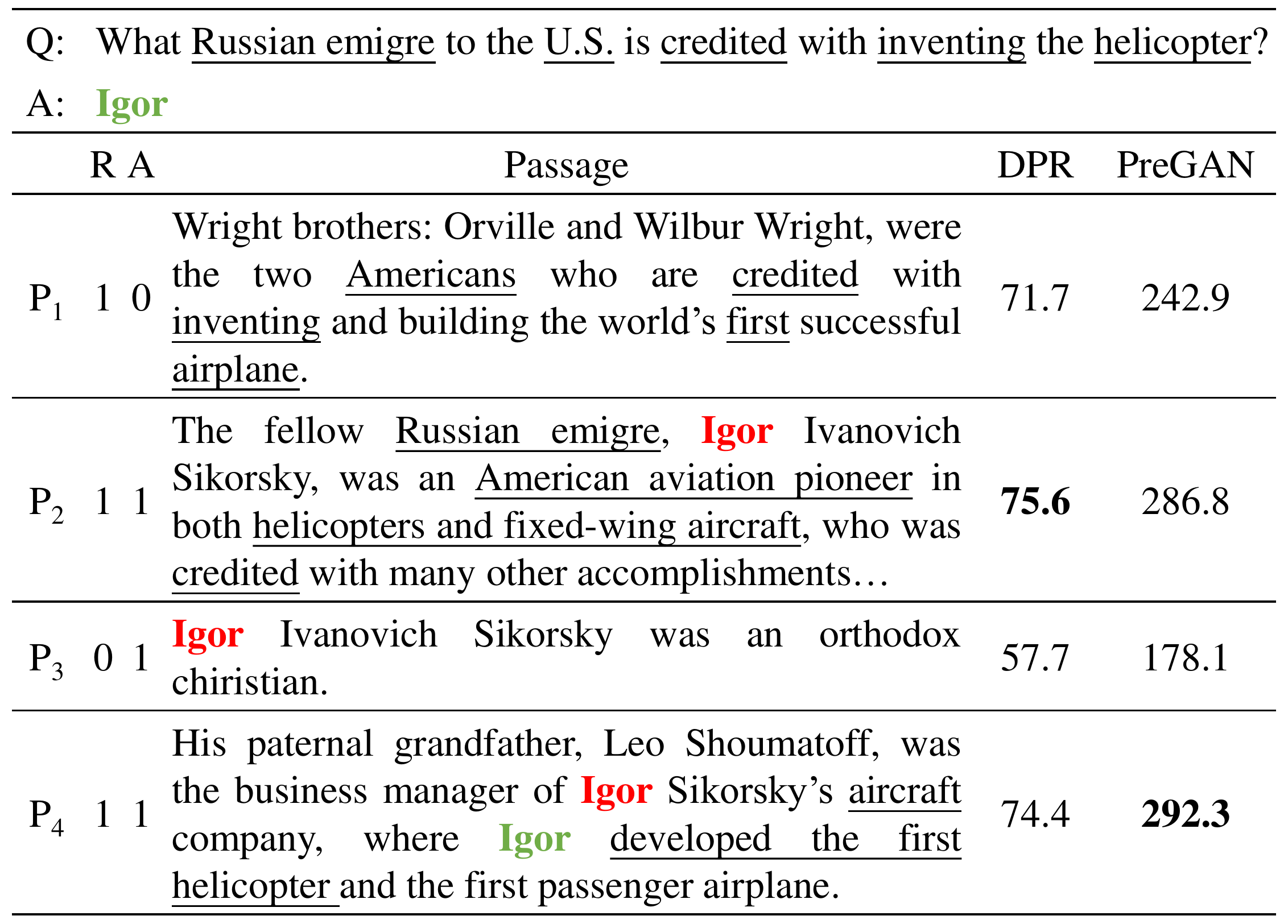}
    \caption{Examples of noisy passages for answering a question from  Quasar-T. R and A indicate if the passage is topically relevant and if it contains the answer, respectively. \underline{Underlined} words are related topic words, and \textcolor[rgb]{0,0.75,0}{green words} are the answer text. \textcolor[rgb]{1,0,0}{Red words} are correct but unsupported answers in the passage. We also show the ranking scores computed by \ttt{DPR} and our \ttt{PReGAN} model.}
    \label{fig:example}
\end{figure}

Training an effective reranking process is not trivial due to the problem of noisy training data: in many cases (datasets), we only know the correct answer to a question, but not the correct passages from which it should be extracted. For example, as illustrated in Figure~\ref{fig:example} (a sample from the Quasar-T dataset), for the question about the inventor of the helicopter, we only know the right answer ``Igor''. We can see that $\mathrm{P_2}$, $\mathrm{P_3}$ and $\mathrm{P_4}$ all contain the answer entity. However, $\mathrm{P_3}$ is obviously not a relevant passage to the question. 
Nevertheless, such a passage has been commonly used as positive training data in previous studies.
Another typical example is $\mathrm{P_1}$, which is highly relevant to the question and contains entities (two persons) that look like the answer.
More tricky cases are $\mathrm{P_2}$ and $\mathrm{P_4}$: Both passages are topically relevant and contain the answer entity. However, $\mathrm{P_4}$ provides support for the answer, while $\mathrm{P_2}$ does not. 
Using all the passages containing the answer as positive examples will obviously confuse the subsequent reader. The existing methods for passage ranking do not perform well on these cases. For example, a recent state-of-the-art QA model \ttt{DPR}~\cite{karpukhin2020dense} ranks $\mathrm{P_2}$ before $\mathrm{P_4}$. Our goal is to rerank the candidates to promote $\mathrm{P_4}$, by enhancing the ranker with some capability of machine reading to detect answerability.


The problem of answer-oriented passage ranking has been addressed in a few recent studies. Two typical approaches have been proposed: incorporating a strong neural mechanism in the first retrieval step so that the question can be naturally expanded (\eg, \ttt{DPR}~\cite{karpukhin2020dense}), or adding a reranking step between the retriever and machine reader (\eg, \ttt{DSQA}~\cite{lin2018denoising,choi2017coarse}). While the dense passage retrieval model (\ttt{DPR}) greatly outperforms the BM25-based retrieval, it is known to require large resources to build the index and to determine the candidate passages, which may not be available in real application situations.
More importantly, it does not incorporate explicitly the criterion of answerability in its retriever. Our approach is more similar to \ttt{DSQA}~\cite{lin2018denoising}, in which a ranker is used between the retriever and the reader. 
However, when training the ranker, none of the previous approaches explicitly distinguished true positives and false positives: 
any passage from which the answer text can be detected (\eg, $\mathrm{P_3}$) was used as a positive training sample for the ranker. 

To better address the problem, we propose the \ttt{PReGAN} model for \textbf{P}assage \textbf{Re}ranking based on \textbf{G}enerative \textbf{A}dversarial \textbf{N}etworks (GAN)~\cite{goodfellow2014generative}, which can force the generator to better assimilate truly positive examples.
We extend the GAN framework by incorporating two discriminators, respectively for topical relevance and for  answerability. 
In so doing, we aim to build a stronger generator, which incorporates some reading capability, to generate passages satisfying both criteria.
In particular, it should assign lower scores to passages of low relevance ($\mathrm{P_3}$), containing incorrect answers ($\mathrm{P_1}$) or not containing support for the answer ($\mathrm{P_2}$), and boost the right passage ($\mathrm{P_4}$), as shown in Figure~\ref{fig:arch}.

The main contributions of this paper are threefold:

(1) We propose a lightweight answer-oriented ranking method  to explicitly model answerability in addition to relevance.

(2) We mitigate the risk of noisy training data     through a customized minimax gaming mechanism.

(3) Experiments on five public OpenQA datasets demonstrate the utility of combining answerability with relevance in passage ranking, which in turn, boosts the quality of extracted answers. 

\section{Related Work}
We only review the most relevant studies to our work in this section. 

Passage ranking is crucial for QA to select better and fewer passages for the reader. 
The naive approach used in many studies relies on an IR model for it~\cite{chen2017reading, yang2019end, yang2019data,nie2019revealing,min2019discrete,wolfson2020break, min2019knowledge, asai2019learning}.
However, an IR model will only rank the passages according to their topical relevance to the question without taking into account the answerability. For example, previous studies~\cite{lin2018denoising} show that a better passage ranking performance on Hits@$N$ may not necessarily lead to a better answer extraction for the whole QA system.
More recent studies have applied dense vector representations to rank passages, based on which a more complex ranking model can naturally incorporate query expansion~\cite{karpukhin2020dense, gao2021scaling, nogueira2019passage,humeau2019poly,khattab2020colbert,das2018multi}. However, the neural retrieval process still focuses on topical relevance only. 

Some recent studies have incorporated an intermediate ranking step on the search results, which considers the feedback from the reader, or relies on some reading capability incorporated into the ranker. 
Typical examples are \ttt{R$^3$}~\cite{wang2018r}, \ttt{DSQA}~\cite{lin2018denoising}, and \ttt{Retro-Reader}~\cite{zhang2020retrospective}. In these approaches, any passage containing the answer text (entity) will be used as a positive sample for ranker training, \ie,
the ranker is not explicitly trained to denoise the training data, and it relies on another component (usually a computationally expensive external reader) to provide indications. The GAN-based approach proposed in this paper specifically addresses the problem of noisy data inside the ranker, so as to simultaneously improve the efficiency and effectiveness of the reader by feeding it with a smaller number of passages with better answerability through ranking.


Generative adversarial networks~\cite{goodfellow2014generative} have been widely used, including in IR~\cite{wang2017irgan,zou2018equilibrium,zhang2018generative} and recommendation~\cite{bharadhwaj2018recgan,cai2018generative,fan2019deep,yuan2020exploring,chong2020hierarchical}. The GAN framework aims at generating the positive samples while making large differences with false positives. In so doing, a stronger generation model can be obtained.
Recently, supervised GAN frameworks have also been explored in  of question answering tasks~\cite{ramakrishnanovercoming,tang2018learning, lewis2018generative, lee2019domain, patro2020robust, joty2017cross, liu2020unified,yang2017semi,oh2019open}, such as community QA, visual QA~\cite{ramakrishnan2018overcoming}, and multi-choice QA, etc. 
GANs are utilized to tackle either the class imbalance problem of data~\cite{yang2019adversarial} or the question clarification problem~\cite{DBLP:journals/corr/abs-1904-02281} in QA. However, none of them explored the utilization of GAN for answer-oriented passage ranking. Our study shows that GAN can be effectively used for this purpose. 

The traditional GAN framework makes a minimax competition between one generator and one discriminator. For passage ranking, we incorporate two discriminators for relevance and answerability. 
Multiple discriminators have been used in some previous GAN frameworks. For example,~\citet{nguyen2017dual} have used a similar approach to tackle the mode collapse problem in GAN. Some others applied similar structure to improve 
image quality~\cite{durugkar2016generative} or for image-to-image translation~\cite{yi2017dualgan}. Our study is the first attempt using such an extended GAN to cope with multiple criteria in passage ranking for open-domain question answering.

There are also several studies that exploit external resources and large pre-trained models for QA ~\cite{lee2019latent,xiong2020progressively,guu2020realm}. The advantages of external resources and pre-trained language models are obvious: the external resources may contain passages that contain the right answer, or the pre-trained language model may already cover the question, \ie, it is able to generate the right answer for it. In these cases, it is difficult to determine where the improvements observed on QA quality come from: from the external resources/pre-trained language model, or the passage ranking strategy using the same source of information. To avoid this confusing situation, we do not include enhancement by external resources or pre-trained language models in this study, although our approach is orthogonal to them and can be combined with them (this will be investigated later).

\section{Methodology}
Our ranking model aims to re-rank the list of passages returned by some efficient retrieval component, such as \ttt{BM25}, so that the passages that can answer the given question will be highly ranked. We can use any reader to read the proposed passages to locate the answers within them. In the following, we only detail the process of reranking. Notice that our ranking model should be efficient so as not to hinder the efficiency of the whole QA process.

\begin{figure}[t!]
    \centering
    \includegraphics[width=.9\linewidth]{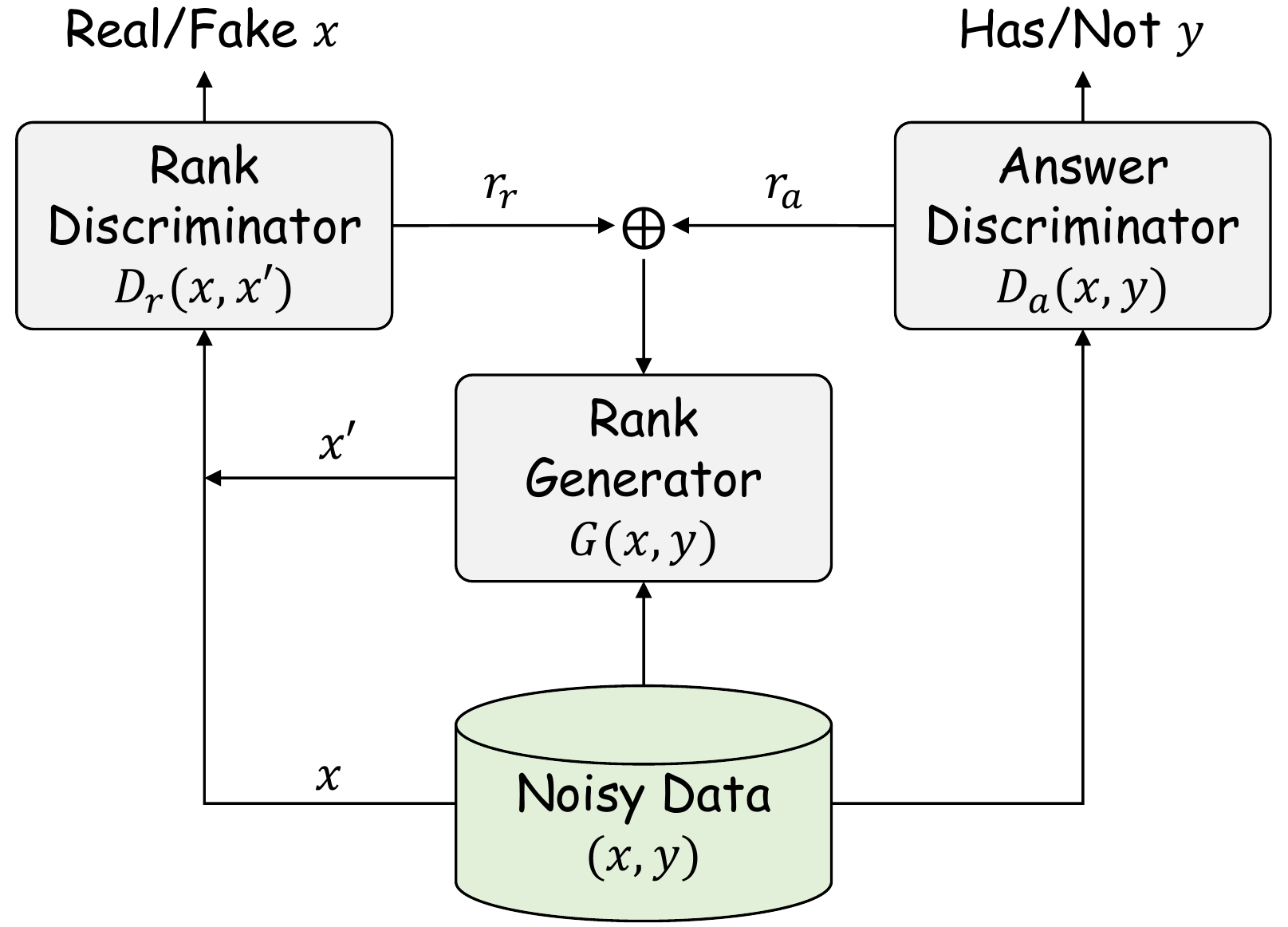}
    \caption{Our framework. $(x, y)$ is a passage-answer pair.}
    \label{fig:arch}
    \vspace{-5pt}
\end{figure}

\subsection{Ranking Framework}
The general architecture of the \ttt{PReGAN} model is illustrated in Figure~\ref{fig:arch}. It is a weakly supervised generative adversarial neural network containing three parts: a rank discriminator (for relevance), an answer discriminator (for answerability), and an answer-oriented rank generator. 

The \textbf{rank discriminator} distinguishes the generated rank distributions from the ground-truth rank distribution. The \textbf{answer discriminator} tells if a generated sample contains the answer to the given question. The \textbf{rank generator} reads the passages under the guidance of the two discriminators and learns to generate the answer-oriented distribution on thoses passages for the given question. The discriminators therefore guide the reading-based generator with a REINFORCE process from two different perspectives. 

At the end of the training, it is expected that passages that are relevant to the question and contain the answer text can be ranked highly. 
In other words, optimizing the three objectives modeled in the GAN framework will help discard noise and promote good  passages that can truly answer the question. 
The utilization of the GAN framework as an effective denoising  means~\cite{tran2020gan} is not new, but we show for the first time that it is also particularly adapted to the noisy training environment in OpenQA.

\subsection{Overall Objective}
Taking the answer-oriented ranking as a distribution over passages for a given question, the objective of the generator is to learn this distribution through the minimax game. The rank generator would try to select passages that the answer can be reasoned (by the reader) for the given question. The rank discriminator would try to draw a clear distinction between the ground-truth answer passages and the selected ones made by the generator. Since the positive training passages are noisy, the rank discriminator alone may be misleading for the machine reading-based generator. However, if the answer can be extracted from the passage and the answer discriminator confirms the correctness of the answer, the  passage selected by the generator should be a good one. This principle can be formulated as the following  overall objective function of the minimax game:
\begin{equation}\label{eq:obj}
    \begin{split}
        J = \min_\theta \max_{\phi,\xi} \sum_{n=1}^N \Big(
        &\mathbb{E}_{d\sim p_{true}(d|q_n,a)} [\log D_\phi^r(d|q_n)] \\
        & + \mathbb{E}_{d\sim p_\theta(d|q_n,a)}[\log (1-D_\phi^r(d|q_n))] \\
        & - \lambda_1\cdot \mathbb{E}_{d\sim p_\theta(d|q_n,a)}[\log D_\xi^a(d|q_n)] \\
        & + \lambda_2\cdot \mathbb{E}_{d\sim p_{true}(d|q_n,a)}[\log \frac{p_{true}(d|q_n,a)}{p_\theta(d|q_n,a)}]
        \Big),
    \end{split}
\end{equation}
where the generative ranking model with parameter $\theta$ is written as $p_\theta(d|q_n,a)$, and the rank discriminator $D_\phi^r$ parameterized by $\phi$ estimates the probability that a passage $d$ can answer the question $q$. The answer discriminator $D_\xi^a$ with parameter $\xi$ works as a regularizer of the generator emphasizing passages in which the answer text  appears, no matter it can answer the question or not. The last item in the objective is another commonly used regularizer~\cite{lin2018denoising,wang2018r} pushing the overall ranking score distribution produced by the generator towards the ground-truth  distribution by minimizing KL-divergence between them. 

\subsection{Rank Discriminator}
As shown in Figure~\ref{fig:arch}, the rank discriminator takes as input the passages generated (selected) by the current optimal generator $p_{\theta^*}$ together with the observed passages with answers, and predicts the source of the passages. Denoting $D_\phi^r(d|q_n) = \sigma(f_\phi(d,q_n))$, which is the sigmoid function of the rank discriminator score $f_\phi$, the loss function of rank discriminator is then represented as:
\begin{equation}\label{eq:dis_r}
    \begin{split}
       \mathcal{L}_{D_\phi^r} =  -\sum_{n=1}^N\Big(&\mathbb{E}_{d\sim p_{true}}[\log(\sigma(f_\phi(d,q_n)))] \\
        &  + \mathbb{E}_{d\sim p_{\theta^*}}[\log(1-\sigma(f_\phi(d,q_n)))]\Big),
    \end{split}
\end{equation}
where the score function $f_\phi$ is implemented as a bi-encoder attention network between passages and questions, and is solved by stochastic gradient descent algorithm. 

\subsection{Answer Discriminator}
The answer discriminator is designed to cooperate with the generator to mitigate the effects of noise in the training data. The answer discriminator only tells if the answer text appears in a passage for the given question instead of trying to reason out a result. The generator, on the other hand, makes decisions based on a machine reading comprehension process and cares less about the existence of the answer texts in the passages.
The idea is that the answer discriminator and generator make judgements from different perspectives, then cross check the judgements and guide the training of the generator towards passages on which they have mutual agreement. The training of the answer discriminator, however, is independent from the generator, and can be trained directly using passage and answer pairs for each question with a binary cross entropy loss  as follows:
\begin{equation}\label{eq:dis_a}
\begin{split}
    \mathcal{L}_{D_\xi^a} = -\sum_{n=1}^N\Big(\sum_{d\in A^+}\log\sigma(f_\xi((d,q_n)))+\sum_{d\in A^-}\log(1-\sigma(f_\xi(d,q_n)))
    \Big), \notag
\end{split}
\end{equation}
where, similar to the rank discriminator, the score function $f_\xi$ is also implemented as a bi-encoder attention network. Passages for question $q_n$ are divided into $A^+$ and $A^-$ depending on whether a passage contains the answer text or not.

\subsection{Generator}
The generator $p_\theta$ aims to minimize the objective, through which it tries to fool the rank discriminator by generating samples faking the ground-truth answer distribution. By keeping the two discriminators $f_{\phi^*}$ and $f_{\xi^*}$ fixed after their maximization, the generator's loss function is as follows: 
\begin{equation}
\begin{split}
    \mathcal{L}_{p_\theta} = \sum_{n=1}^N\Big(~&\mathbb{E}_{d\sim p_\theta}[\log(1-\sigma(f_{\phi^*}(d,q_i)))] \\
    & -\lambda_1~ \mathbb{E}_{d\sim p_\theta}[\log \sigma(f_{\xi^*}(d,q_n))] \\
    & -\lambda_2~ \mathbb{E}_{d\sim p_{true}}[\log p_\theta(d|q_n,a)]\Big),
\end{split}
\end{equation}
where minimizing the first item means to guide the generator $p_\theta$ toward generating passages which the rank discriminator considers as likely ground-truth passages. Minimizing the second item will drive the generator away from  generating passages that the answer discriminator dislikes. The third item imposes a regularization on the overall answer distribution produced by the generator, forcing it to stay close to the ground-truth answer distribution. $\lambda_1$ and $\lambda_2$ are hyper-parameters controlling the impacts of the answer discriminator and the distribution regularizer on the generator.

The generator is trained under weak supervision using the noisy training data. It predicts the ranking score of each passage using a simple machine reading comprehension (MRC) component. Specifically, the MRC component takes a question and the passage candidates as input, and produces a joint distribution of the start position and end position of the answer in a passage. Then it takes the maximum of the joint probability as the predicted ranking score of the passage with respect to the question. The ranking scores are then normalized into a probabilistic distribution indicating how likely each passage can answer the given question. Similar to $f_\phi$ and $f_\xi$, the ranking scores are also generated by scoring networks. The details of the scoring networks are given in the next section.
Our scoring networks are inspired by~\cite{lin2018denoising}. We opt for it because of its high efficiency, which is required for our ranking. Other scoring schemes, such as transformer-based approaches, could also be adopted in our framework in the future. 

From the loss function of the generator, its gradient can be derived as follows (We include the details about the derivation of Equation~(\ref{eq:policy}) in the Appendix~\ref{sec:reward} for interested readers):
\begin{equation}\label{eq:policy}
    \begin{split}
        \nabla_\theta~ \mathcal{L}^G (q_n) = & \nabla_\theta \mathbb{E}_{d\sim p_\theta}\left[\log(1+\exp(f_\phi(d,q_n)))\right] \\
        & +\lambda_1 \nabla_\theta \mathbb{E}_{d\sim p_\theta}\left[\log(1+\exp(f_\xi(d,q_n)))\right] \\
        & -\lambda_2 \nabla_\theta \mathbb{E}_{d\sim p_{true}}[\log p_\theta(d|q_n, a)]\\
        = & \frac{1}{K} \sum_{k=1}^K\nabla_\theta \log p_\theta(d_k|q_n,a)\Big(\log(1+\exp(f_\phi(d_k,q_n)))\\ 
        & + \lambda_1 \log(1+\exp(f_\xi(d_k,q_n))) \Big) \\
        & -\lambda_2 \sum_{m=1}^M \frac{p_{true}(d_m|q_n,a)}{ p_\theta(d_m|q_n, a)},
    \end{split}
\end{equation}
where $p_{true}=1/|d^+|$ ($|d^+|$ is the total number of passages containing correct answers) if the passage contains a correct answer, otherwise 0. This completes the approximation approach for generator optimization with policy gradient based reinforcement learning. The term $\log(1+\exp(f_\phi(q_n,a))) + \lambda_1 \log(1+\exp(f_\xi(q_n,a)))$ acts as the balanced rewards from rank discriminator and answer discriminator for the generator $p_\theta(d|q_n,a)$ to select passage $d$ for a given question $q_n$.

\subsection{Scoring Networks}\label{sec:networks}
In this section, we will introduce the scoring functions we used in details. As can be seen from the Figure~\ref{fig:arch} and the Equation~(\ref{eq:obj}), the three scoring functions lie in the kernel of our proposed GAN framework.

Formally, given a question $q=(q^1, q^2, \cdots, q^{|q|})$ and $m$ passages returned by some retrieval component, which are defined as $D=\{d_1, d_2, \cdots, d_m\}$, where $d_i=(d_i^1, d_i^2,\cdots,d_i^{|d_i|})$ is the $i$-th retrieved passage composed of a sequence of words, the answer-oriented passage ranking aims to assign a score $P(d_i|q)$ to passage $d_i$ measuring the probability that it can answer the question $q$.

For $\mathbf{f_\phi}$ \textbf{in rank discriminator $D_r$}, we use a bi-encoder framework to build the scoring function $f_\phi(d,q_n)$. Each passage $d_i$ and the question $q$ are encoded with an RNN network:
\begin{equation*}
\small
    \begin{split}
        <\hat{d}_i^1,\hat{d}_i^2,\cdots,\hat{d}_i^{|d_i|}>~&=~\text{RNN}(<d_i^1,d_i^2,\cdots,d_i^{|d_i|}>),\\
        <\hat{q}^1,\hat{q}^2,\cdots,\hat{q}^{|q|}>~&=~\text{RNN}(<q^1,q^2,\cdots,q^{|q|}>),
    \end{split}
\end{equation*}
where $\hat{d}_i^j$ and $\hat{q}^j$ are the hidden representations of words in passage $d_i$ and question $q$ encoding its context information through RNN. We use a single-layer bidirectional LSTM (biLSTM) as our RNN unit. Following~\cite{lin2018denoising}, the final representation of the question is obtained through a self-attention operation $\hat{q} = \sum_{j=1}^{|q|}\alpha^j\hat{q}^j$, where $\alpha^j=\text{softmax}(wq^j)$ is the importance of each word in the question and $w$ is a learned weight vector. Then the score of each passage is obtained via a max-pooling layer and a softmax layer:
\begin{equation*}
    f_\phi(d_i, q) = p(d_i|q) = \text{softmax}(\max_j(\hat{d}_i^jWq)),
\end{equation*}
where $W$ is a weight matrix to be learned.

For $\mathbf{f_\xi}$ \textbf{in answer discriminator $D_a$}, even though it serves a different purpose with a different objective function from the rank discriminator $D_r$, the network framework is similar, except that the last layer is a sigmoid function instead of softmax function.

Finally, for  $\mathbf{f_\theta}$ \textbf{in rank generator $G$}, similar to $\mathbf{f_\phi}$, we encode the passage $d_i$ in a sequence of hidden vectors, and apply a self-attention layer to attend to the question vectors to get a question embedding. Different from the discriminators, we ask the generator to score a passage through predicting the probability of the start and end positions of an answer span in the passage. The probability of a position $j$ as the start position of the answer span in the passage $d_i$ is calculated by $p_s^j = \text{softmax}(\hat{d}_i^jW_s\hat{q})$ with parameter $W_s$, and $p_e^j = \text{softmax}(\hat{d}_i^jW_e\hat{q})$ for the end position $p_e^j$ using $W_e$. The score of passage $d_i$ is then obtained as the maximum of the joint probability of the start and end positions:
\begin{equation*}
\small
    f_\theta(d_i, q) = p_\theta(a|q,d_i) = \max_{j,k} p_s^j(a|q,d_i)p_e^k(a|q,d_i) 
\end{equation*}

Our scoring functions are inspired by~\cite{lin2018denoising}. We opt for it because of its high efficiency, which is required for our ranking. Other scoring schemes, such as transformer-based approaches, could also be adopted in our framework in the future.

\subsection{Ranking Algorithm}\label{sec:algo}
With all the components in place, the overall ranking algorithm is summarized in Algorithm~\ref{alg:game}.
Before the adversarial training, we initialize the three component with random parameter weights. The answer discriminator, rank discriminator, and generator are pre-trained using the noisy training data. 

From the ranked passages, the top-K (K=50 in our experiments) are passed to the subsequent reader to determine the answer span. 

We use a \ttt{BERT}-based reader architecture in this paper, which follows~\cite{karpukhin2020dense}.\footnote{\url{https://github.com/facebookresearch/DPR}}
Let $d_i\in \mathbb{R}^{L\times h}$ ($i\leq i \leq k$) be a BERT representation of the $i$-th passage, where $L$ is the maximum length of the passage and $h$ is the hidden dimension.\footnote{We assume that the reader is familiar with the BERT training process, or could find more details in~\cite{karpukhin2020dense}.}  The probability of an answer span ($s,e$) in passage $d_i$ is defined as:
\begin{align*}
    P(s,e,i) &= P(d_i) \cdot P(s|d_i) \cdot P(e|d_i), \\
    P(s|d_i) &= \text{softmax}(d_i w_{start})_s, \\
    P(e|d_i) &= \text{softmax}(d_i w_{end})_t, \\
    P(d_i) &= \text{softmax}(\hat{D}^T w_{doc})_i,
\end{align*}
where $\hat{D} = [d_1^{[CLS]},\cdots,d_k^{[CLS]}]\in \mathbb{R}^{h\times k}$ where $d_i^{[CLS]}$ is the \ttt{BERT}-based sentence representation (with 12 transformer layers and hidden size of 768), and $w_{start}$, $w_{end}$, $w_{doc}\in \mathbb{R}^h$ are learnable parameters. 
Notice that the final answer's probability
$P(s,e,i)$ also uses the passage ranking score $P(d_i)$, so that the reader will trust more the answers from highly ranked passages. 

\begin{algorithm}[t]
\caption{Answer-Oriented Ranking Algorithm}
 \label{alg:game}
\textbf{Input}: generator $f_\theta$, rank discriminator $f_\phi$, answer discriminator $f_\xi$, training dataset $D$
\begin{algorithmic}[1]
 \STATE Initialize $f_\theta$, $f_\phi$, $f_\xi$ with random weights $\theta, \phi, \xi$, num-epochs, d-steps, g-steps.
 \STATE Pre-train the rank discriminator $f_\phi$, answer discriminator $f_\xi$, and the generator $f_\theta$ using $D$.
 \WHILE{num-epochs}
 \STATE num-epochs~$--$
  \WHILE{g-steps~$>0$}
    \STATE g-steps~$--$
    \STATE Sample $K$ passages for each query $q$ by $f_\theta$;
    \STATE Collect rewards from answer discriminators $f_\xi$ for the $K$ passages;
    \STATE Collect rewards from rank discriminators $f_\phi$ for the $K$ passages;
    \STATE Update the generator $f_\theta$ via policy gradient Equation~(\ref{eq:policy}).
  \ENDWHILE
  \WHILE{d-steps~$>0$}
    \STATE d-steps~$--$
    \STATE Sample $K$ passages for each query $q$ by $f_\theta$ as negative passages;
    \STATE Combine negative passages with positive passages in $D$ for each question $q$;
    \STATE Update the rank discriminator $f_\phi$ via Equation~(\ref{eq:dis_r}).
  \ENDWHILE
 \ENDWHILE
 \end{algorithmic}
\end{algorithm}

\section{Experiments}

\subsection{Datasets and Evaluation Metrics}
We evaluate our ranking model on five commonly-used datasets for OpenQA tasks:

\textbf{Quasar-T}~\cite{dhingra2017quasar} contains 43,012 trivia questions, each with 100 passages retrieved from ClueWeb09 data source using LUCENE. 

\textbf{SearchQA}~\cite{dunn2017searchqa} is a large-scale OpenQA dataset with question-answer pairs crawled from J! Archive and passages retrieved by Google. It contains 140k question-answer pairs. On average, each question has 49.6 passages.

\textbf{TriviaQA}~\cite{joshi2017triviaqa} includes 95K question answer pairs authored by trivia enthusiasts and independently gathered evidence passages. Each question has 100 webpages retrieved by Bing Search API. We only use the first 50 for training the ranker and the reader.

\textbf{Curated TREC}~\cite{baudivs2015modeling} is based on the benchmark from the TREC QA tasks, which contains 2,180 questions extracted from datasets of TREC 1999, 2000, 2001, and 2002.


\textbf{Natural Questions}~\cite{kwiatkowski2019natural} consists of real anonymized, aggregated questions issued to the Google search engine. The answers are spans in Wikipedia articles identified by annotators. Each question is paired with up to five reference answers. 

For Quasar-T, SearchQA, and TriviaQA, we use the same processed paragraphs provided by~\cite{lin2018denoising}. For Curated TREC and Natural Questions, following existing work, we determine a subset of 50 passages for each question and apply our model to them~\cite{lin2018denoising,min2021neurips}.~\footnote{This is done due to our limited computation resources -- training our model on a larger set of long passages (webpages) would require more memory than we have.} We call the generated datasets ``Natural Questions Subset'' (NQ-Sub).
The statistics are shown in Table~\ref{tab:data}.
\begin{table}[t]
    \centering
    \small
    \caption{Statistics of the datasets.}
    \begin{tabular}{lrrrc}
    \toprule
         \textbf{Dataset} & \textbf{Train} & \textbf{Dev.} & \textbf{Test} &
         \textbf{\#Passages/Question}\\
    \midrule
         Quasar-T & 37,012 & 3,000 & 3,000 & 100 \\
         Search QA & 99,811 & 13,893 & 27,247 & $\sim$49.6   \\
         Trivia QA & 87,291 & 11,274 & 10,790 & 100   \\
         Curated TREC & 1,353 & 133 & 694 & Wiki (50) \\
         Natural Question & 79,168 & 8,757 & 3,610 & Wiki (50)\\
    \bottomrule
    \end{tabular}
    \label{tab:data}
\end{table}

The ranking results are evaluated by two common metrics:

\textbf{HITS@$N$} evaluates the ranking results by indicating the proportion of top-$N$ passages that contains answer texts. Notice that this measure only provides an approximation of the true ranking quality, as a passage containing the answer without support would also be considered relevant.

\textbf{EM} measures the percentage of answers found by the whole system (the reader) that match  the ground-truth answers. 
\subsection{Baselines}
For comparison, we select four representative ranking-based models as baselines. These models do not exploit external resources. 

(1) \ttt{BM25+BERT}: The passages ranked by \ttt{BM25} are directly submitted to the \ttt{BERT}-based reader. This is intended to test how a basic retriever+reader approach could perform on the datasets. 

(2) \ttt{DPR}~\cite{karpukhin2020dense} is one of the latest methods proposed in the literature, and it produces state-of-the-art performance. It adopts a bi-step retriever-reader framework where the retriever utilizes a neural matching model. The reader is the same \ttt{BERT}-based network structure as ours. We will note that \ttt{DPR} requires large computation resources to run.

(3) \ttt{DSQA}~\cite{lin2018denoising} adopts a  retriever-selector-reader framework similar to ours. However, its denoising is limited to exploiting the score of the reader as a confidence score. No explicit denoising is done within the selector. 
The comparison with \ttt{DSQA} will reveal the impact of the GAN framework for denoising in ranking. 

(4) \ttt{R$^3$}~\cite{wang2018r} is a reinforced model using  a ranker to select the most confident passage to pass to the reader. Interactions are allowed between the ranker and the reader -- the reader provides feedback (rewards) to the ranker. This is similar to the principle of our ranker, but we use GAN to discriminate between different passages and force the ranker to meet multiple objectives. In addition, the same neural architecture is used in \ttt{R$^3$}'s reader and ranker, while we employ a simplified reader in our ranker for higher efficiency.

For fairness, we do not include  methods such as \ttt{REALM}~\cite{guu2020realm} and \ttt{ORQA}~\cite{lee2019latent} which either utilize external resources or expensive additional pre-training tasks, while our method does not. These approaches are orthogonal to ours, and can be added on top of ours. We leave it to future work.

\subsection{Implementation Details}
We tune our model on the development set. Restricted by the computation resources available -- one Nvidia GTX Titan X GPU (12G VRAM), only limited parameter sets could be explored. The hidden size of the RNN used in the ranker is {128}, the number of RNN layers for the passage and question are both set to {1}. The maximum passage length is set according to the passage length distributions -- 150 for quasar-T, SearchQA, TriviaQA, and Natural Questions, and 350 for Curated TREC. The batch size varies among \{2, 4, 8, 16\} across datasets depending on the maximum passage length in order to fit the model into the GPU RAM. We only use top-50 passages retrieved by BM25 
to train the ranker and the reader. The hyper-parameters $\lambda_1$ and $\lambda_2$ are set to 0.25 and 1. We  follow ~\cite{lin2018denoising}  for the setting of other parameters. The complete optimal parameter settings and code will be made public in our Github repository later.

\subsection{Ranking Results}
We first test the ability of our \ttt{PReGAN} ranker to better rank passages containing the answers. The baseline models have been used on different datasets. To compare with them directly, we run two sets of experiments: one on Quasar-T and SearchQA on which \ttt{DSQA} has been tested, and another on Curated TREC and TriviaQA on which \ttt{DPR} has been tested. In addition, we also compare \ttt{DPR} with our method on the Natural Question subset.

At the top of Table~\ref{tab:baselines}, we report the ranking results by \ttt{BM25}, \ttt{DSQA}, and \ttt{PReGAN}. It shows that \ttt{PReGAN} outperforms both baselines by a large margin. 
Recall that the main difference between \ttt{DSQA}'s selector and \ttt{PReGAN} lies in the denoising based on GAN. The improvements hence directly reflect the usefulness of the GAN mechanism.

\begin{table}[t]
    \centering
    \small
    \caption{Ranker performance compared with \ttt{DSQA} on datasets Quasar-T and SearchQA, and compared with \ttt{DPR} on Curated TREC, TriviaQA, and NQ-Sub. The best results are in bold. Results marked with $\Diamond$ are cited from~\cite{lin2018denoising}, which does not provide Hits@20 and Hits@50. Results marked with $\triangle$ are cited from~\cite{karpukhin2020dense}. Results marked with $\heartsuit$ are obtained by running \ttt{DPR} against top-50 passages returned by \ttt{BM25}.}
    \begin{tabular}{lccccc}
    \toprule
        \textbf{Quasar-T} & Hits@1 & Hits@3 & Hits@5 & Hits@20 & Hits@50 \\
    \midrule
        \ttt{BM25}$^\Diamond$ & 6.3 & 10.9 & 15.2 & - & - \\
        \ttt{DSQA}$^\Diamond$ & 27.7 & 36.8 & 42.6 & - & - \\
        \ttt{PreGAN} & \textbf{35.2} & \textbf{52.0} & \textbf{59.5} & \textbf{72.3} & \textbf{74.8} \\
    \midrule
        \textbf{SearchQA} & Hits@1 & Hits@3 & Hits@5 & Hits@20 & Hits@50 \\
    \midrule
        \ttt{BM25}$^\Diamond$ & {13.7} & {24.1} & {32.7} & - & - \\
        \ttt{DSQA}$^\Diamond$ & 59.9 & 69.8 & 75.5 & - & - \\
        \ttt{PreGAN} & \textbf{63.9} & \textbf{83.0} & \textbf{88.8} & \textbf{97.5} & \textbf{99.8} \\
    \midrule
        \textbf{Curated TREC} & Hits@1 & Hits@3 & Hits@5 & Hits@20 & Hits@50 \\
    \midrule
        \ttt{BM25} & 23.9 & 45.1 & 54.3 & 73.7 & 82.8 \\
        \ttt{DPR}$^\triangle$ & - & - & - & 79.8 & - \\
        \ttt{PreGAN (50)} & 30.3 & 51.2 & 60.8 & 78.6 & 82.8 \\
        \ttt{PreGAN (100)} & 33.3 & 54.6 & 64.1 & 81.0 & 84.6 \\
    \midrule
        \textbf{TriviaQA} & Hits@1 & Hits@3 & Hits@5 & Hits@20 & Hits@50 \\
    \midrule
        \ttt{BM25} & 28.1 & 46.7 & 56.6 & 75.6 & 81.3 \\
        \ttt{DPR}$^\triangle$ & - & - & - & 79.4 & - \\
        \ttt{PreGAN (50)} & 48.9 & 63.8 & 70.0 & 79.2 & 81.5 \\
        \ttt{PreGAN (100)} & 48.5 & 64.1 & 70.0 & 79.2 & 81.7 \\
    \midrule
        \textbf{NQ-Sub} & Hits@1 & Hits@3 & Hits@5 & Hits@20 & Hits@50 \\
    \midrule
        \ttt{BM25} & 17.6 & 28.9 & 35.5 & 53.1 & 64.3 \\
        \ttt{DPR (all)} & 39.3 & 53.6 & 58.8 & 68.9 & 73.7 \\
        \ttt{DPR (50)}$^\heartsuit$ & 23.6 & 34.8 & 42.4 & 59.7 & 64.3 \\
        \ttt{PreGAN} & 24.0 & 36.7 & 43.3 & 58.2 & 64.3 \\
    \bottomrule
    \end{tabular}
    \label{tab:baselines}
\end{table}

On Curated TREC and TriviaQA, as the candidate passages are not provided in the datasets, but retrieved from Wikipedia, we consider two cases -- retrieving 50 and 100 candidates with \ttt{BM25} and submitting them to the ranker. Notice that \ttt{DPR} directly retrieves passages from the whole Wikipedia dump, and thus is not limited by the 50 or 100 candidate passages. Nevertheless, we can see that our ranker can produce similar ranking results (Hits@20) to \ttt{DPR} despite the more limited candidates. On the Natural Question subset, we run \ttt{DPR} retriever and our ranker on the top-50 results from \ttt{BM25}. We see that \ttt{PReGAN} can better rank passages in top positions than \ttt{DPR}. However, at lower rank positions (from HITS@20), \ttt{DPR} performs better. This can be attributed to the more sophisticated retrieval model in \ttt{DPR}. In the future, the score networks used in our current model could be replaced by more sophisticated ones as in \ttt{DPR}. Another factor that explains the better ranking for \ttt{DPR} when $N \ge 20$ is that HITS@$N$ is only an approximation of the ranking quality. This measure should be considered together with the EM measure of answers.

The most important observation to make is the improvements of \ttt{PReGAN} over \ttt{BM25}. They reflect the capability of our ranker to favor the passages that may contain an answer. We expect that this will benefit the reader, as we will show.

\subsection{Final Answer Results}

We adopt a \ttt{BERT}-based reader structure as in~\cite{karpukhin2020dense} to extract answers from the selected passages. We test the QA answer for each question produced by the reader with the top-50 passages (together with their ranking scores). The results are shown in Table~\ref{tab:overall}. 

\begin{table}[t!]
    \centering
    \small
    \caption{Answer results with exact matching (EM) rate. The best results are in bold. Numbers with $*$ for \ttt{DSQA} and \ttt{R$^3$} are cited from~\cite{lin2018denoising}, those for \ttt{DPR} are from~\cite{karpukhin2020dense}. For the Natural Question subset, our model is based on the top-50 passages retrieved by \ttt{BM25}, while \ttt{DPR} uses its own retriever to retrieve top-50 passages. 
   }
   \setlength{\tabcolsep}{0.9mm}{
    \begin{tabular}{lccccc}
    \toprule
        & \textbf{Quasar-T} & \textbf{SearchQA} & \textbf{Curated Trec} & \textbf{TriviaQA} & \textbf{NQ-Sub}
         \\
    \midrule
        \ttt{BM25+BERT} & 41.6 & 57.9 & {21.3}$^*$ & {47.1}$^*$  & 26.7 \\
        \ttt{R$^3$} & {35.3}$^*$ & {49.0}$^*$ & {28.4}$^*$ & {47.3}$^*$ & - \\
        \ttt{DSQA} & {42.2}$^*$ & {58.8}$^*$ & {29.1}$^*$ & {48.7}$^*$ & -  \\
        \ttt{DPR} & - & - & {28.0}$^*$ & {57.0}$^*$ & 27.4 \\ 
        \ttt{PReGAN} & \textbf{45.5} & \textbf{61.2} & \textbf{29.3} & \textbf{60.7} & \textbf{29.5}  \\
    \bottomrule
    \end{tabular}
    }
    \label{tab:overall}
\end{table}

We can observe that on all the datasets, \ttt{PReGAN} (combined with \ttt{DPR} reader) leads to consistently better answer results than the baselines. 
As \ttt{BM25} ranking results are submitted to the same reader as \ttt{PReGAN}, the differences between them are directly attributed to passage ranking. This comparison clearly shows the usefulness of adding a ranker to rerank the \ttt{BM25} retrieval results. 

The comparison with \ttt{R$^3$} shows the utility of adding discriminators. Recall that \ttt{R$^3$} also uses the rewards from the reader to rerank the retrieval results, but without contrasting between different passages. We can see that by adding discriminators, \ttt{PReGAN} is forced to better distinguish between truly good passages from  noisy ones.

In this experiment, \ttt{DSQA} uses its own selector to rerank the \ttt{BM25} candidates and its reader to find the answer, and \ttt{DPR} also uses both its retriever to retrieve candidates and its reader. This setting may play in favor of these models because the retriever, selector, and reader have been optimized together. More particularly, \ttt{DPR} is allowed to search for top-ranked candidate passages directly from the document collections in Curated TREC (top-100) and Natural Questions (top-50), instead of using the top-50 retrieved with \ttt{BM25}. The quality of these passages is better than that of \ttt{BM25}. Despite this, compared to \ttt{DSQA} and \ttt{DPR}, \ttt{PReGAN} can still produce better results. This result demonstrates that our QA approach can be competitive against the state-of-the-art approaches that have much more parameters. Taking the experiments on passage ranking and answer identification together, we can see that a better passage ranking generally leads to better answers. 

\subsection{Further Analysis}
In this section, we further analyze the impact of the components of \ttt{PReGAN} and the time efficiency.

\subsubsection{{Influence of the Number of Retrieved Passages}}
We test with different numbers of initial retrieval results on Curated TREC, on which we can retrieve the desired number of passages. In the experiment, our ranker is asked to select the top-50 passages from the initial retrieval list of variable sizes.

\begin{figure}[t!]
    \centering
    \includegraphics[width=.9\linewidth]{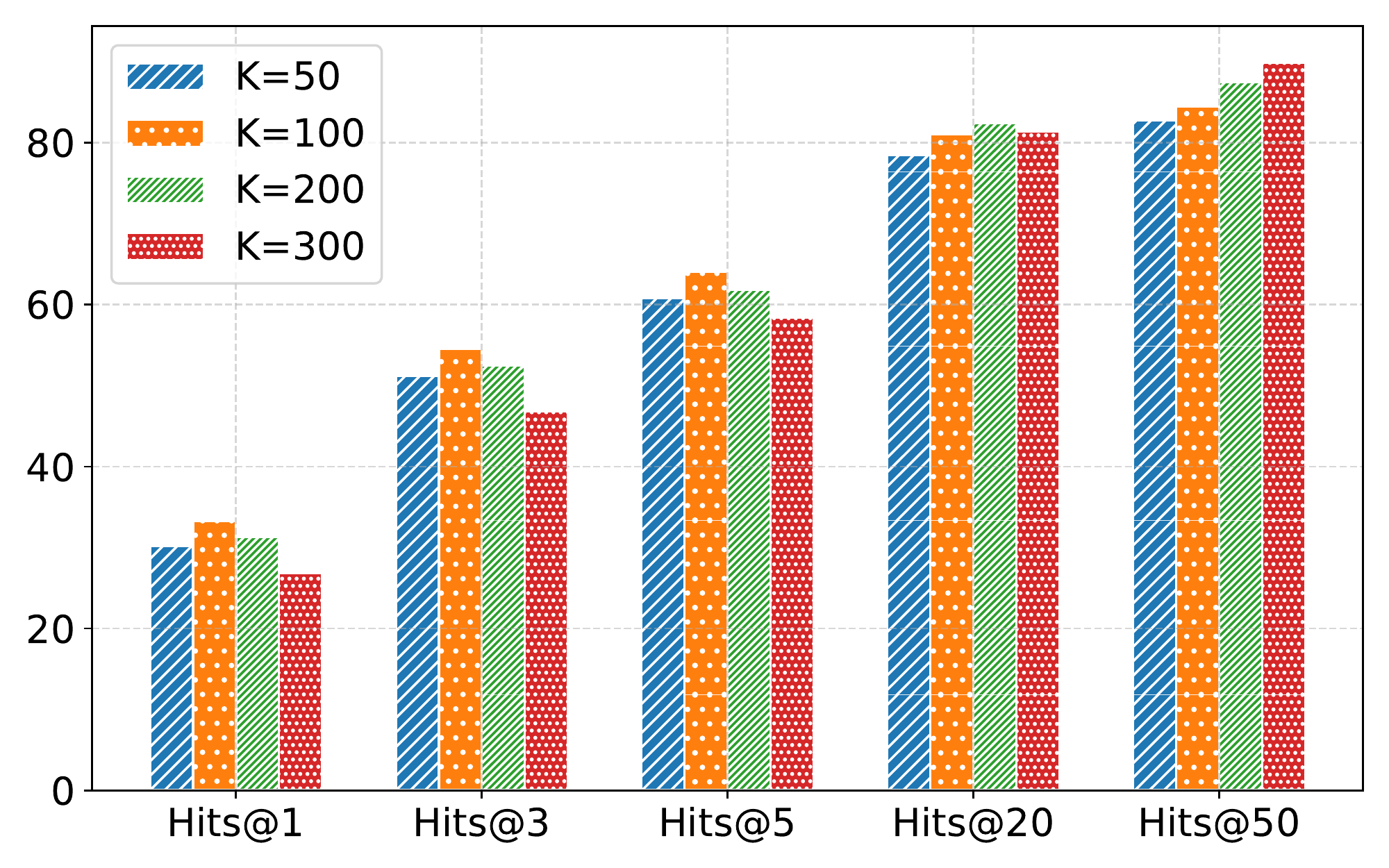}
    \caption{Ranker performance with different numbers of retrieved passages on Curated Trec. }
    \label{fig:trec}
\end{figure}

Figure~\ref{fig:trec} shows how the HITS measures vary depending on the size of the retrieved passages (from 50 to 300).
The general observation is that by increasing the size of initial retrieval results from 50 to 100, all HITS measures are improved. However, when we use a larger set of retrieval results, HITS at small $N$ are hurt, while HITS at larger $N$ keep being improved. This observation tends to suggest that the initial retrieval results could be increased to a reasonable size. However, a too large set of initial retrieval results may lead to a higher likelihood to include noise passages, which may make it more difficult for the ranker to determine the right passages. The right choice of the size of initial passages retrieval is an interesting question that we will investigate in the future.


\subsubsection{Impact of the Number of Passages to be Read}
We select top-$K$ passages from the ranker, and submit them to the reader to find an answer. We want to test the impact of $K$ on the answer results. 

\begin{figure}[t!]
    \centering
    \includegraphics[width=.9\linewidth]{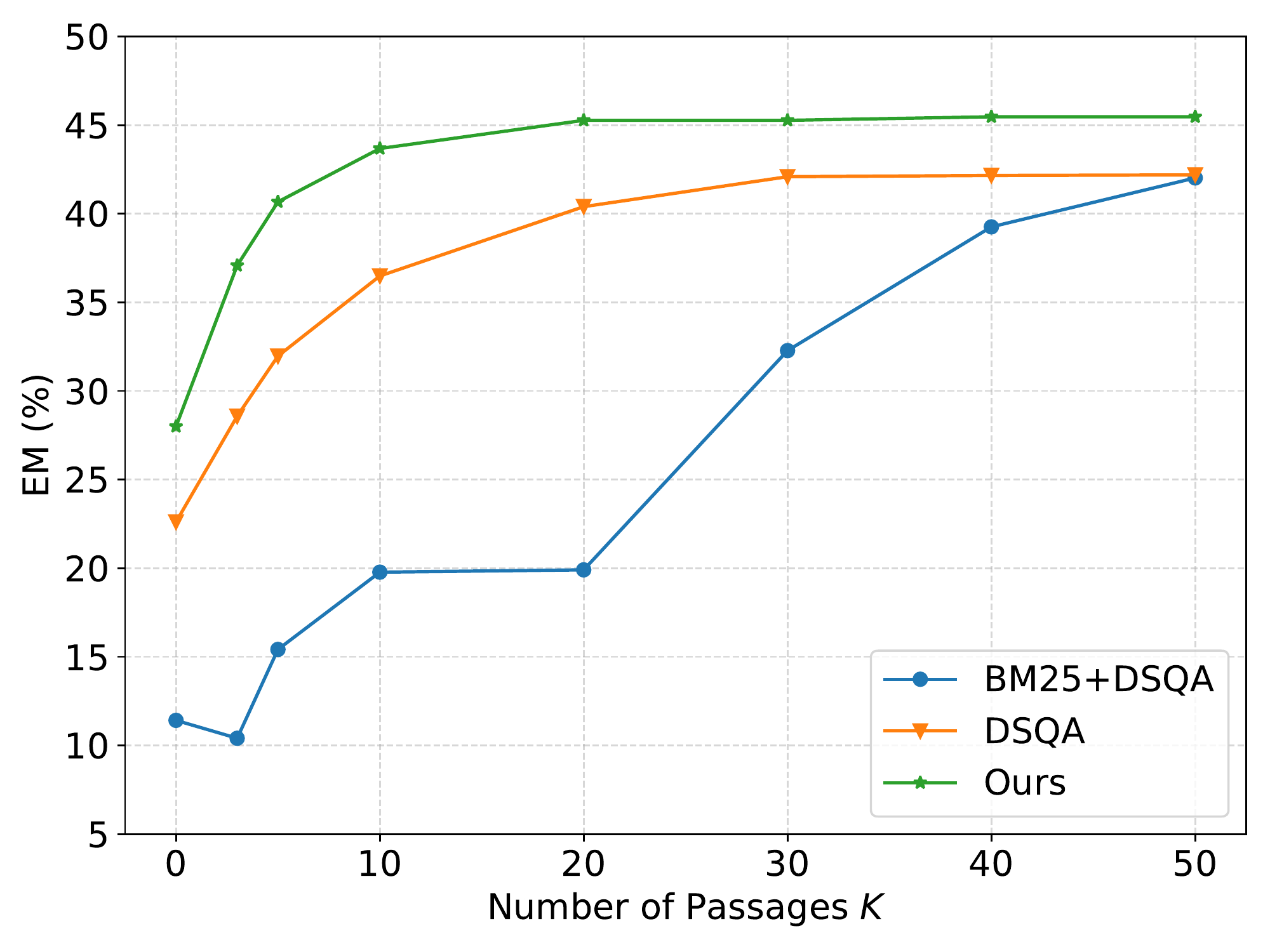}
    \includegraphics[width=.9\linewidth]{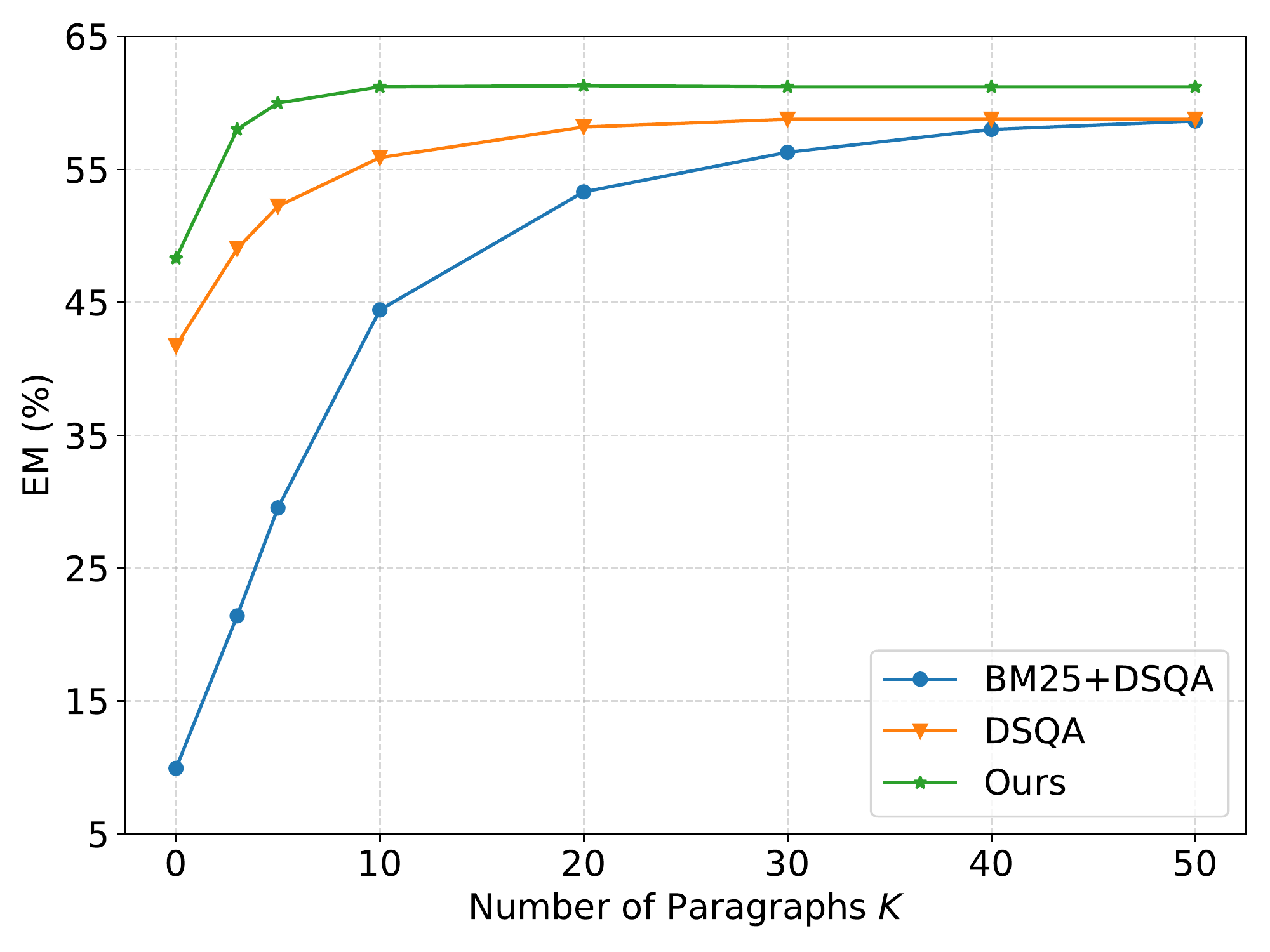}
    \caption{Reader performance with different numbers of ranked passages on Quasar-T (top) and SearchQA (bottom).}
    \label{topk:a}
\end{figure}

Figures~\ref{topk:a} shows the Exact Match results of our system compared with the baseline methods \ttt{DSQA} and \ttt{BM25+DSQA} on the datasets Quasar-T and SearchQA. In \ttt{BM25+DSQA}, the passages are ranked by \ttt{BM25} scores, and we use the reader of \ttt{DSQA} to find the answer.

We can observe that submitting more passages to the reader will generally lead to better answers. However, this is at the cost of higher time complexity due to the complex machine reading process. A good ranker should be able to select a small number of passages for the reader, without penalizing the final results. This is what we observe from \ttt{PReGAN}: by selecting top-20 and top-10 passages for the reader, our model can already obtain the best answer results. This further confirms that \ttt{PReGAN} is able to rank good passages on top. We attribute this capability to the GAN mechanism used, which helps to discriminate between good and bad passages.

Note that \ttt{BM25+DSQA} and \ttt{DSQA} use the same reader and they differ only in an additional selector in \ttt{DSQA}. \ttt{PReGAN} uses a different reader.
To see the contribution of the ranker on a fair ground, we will run another experiment using the same reader. 

\subsubsection{{Effectiveness of Ranker}} In this experiment, we test the respective contributions of the ranker and the reader. We compare our method with \ttt{DSQA}. To test the effectiveness of the ranker, we replace the ranker in \ttt{DSQA} by that of \ttt{PReGAN}, but still use the same \ttt{DSQA} reader to find the answer (\ttt{PReGAN+DSQA}). Table~\ref{tab:DSQA_reader} shows the results.

\begin{table}[t]
    \small
    \centering
    \caption{EM performance of Bi-LSTM-based reader on our ranking results on Quasar-T.}
    \begin{tabular}{lcccc}
    \toprule
         Number of Passages ($k$) & 1 & 3 & 5 & 10 \\\midrule
         \ttt{DSQA} & 22.6 & 28.6 & 32.0 & 36.5 \\
         \ttt{PReGAN+DSQA} & {25.9} & {31.2} & {34.7} & {38.8} \\
         \ttt{PReGAN} (\ttt{DPR} reader) & \textbf{28.0} & \textbf{37.1} & \textbf{40.7} & \textbf{43.7} \\
    \bottomrule
    \end{tabular}
    \label{tab:DSQA_reader}
\end{table}

We can see that when the \ttt{DSQA} ranker is replaced by ours in \ttt{PReGAN+DSQA}, the EM measures are improved (over \ttt{DSQA}). This is a clear demonstration that our ranker \ttt{PReGAN} is more effective than that of \ttt{DSQA}. The comparison between our method and \ttt{PReGAN+DSQA} shows the impact of a better reader -- both our method and \ttt{PReGAN-DSQA} use the same list of passages, but \ttt{PReGAN} is followed by a \ttt{BERT}-based reader (similar to \ttt{DPR}), while \ttt{DSQA} uses a biLSTM-based reader. This shows that the quality of both the ranker and the reader contribute to the global effectiveness of QA, and they are complementary.

\subsubsection{{Time Efficiency}} As an intermediate step, a ranker should be efficient. To show the time cost of different steps, we report in Table~\ref{tab:time_on_nqsub} the time required for different steps on $\text{NQ}_\text{sub}$ on which we can perform all the retrieval-ranking-reading processes.
The experiment is run on a server equipped with
one Nvidia GTX Titan X GPU (12GB VRAM), one Intel Core i7@3.7GHz CPU with 12 cores, and 32G RAM size.

\begin{table}[t!]
\small
\centering
\caption{Total time of retriever, ranker, and reader on all questions in NQ-Sub, and the average time per question.}
\begin{tabular}{lcccc}
\toprule
& \multicolumn{2}{c}{\textbf{Total (s)}} & \multicolumn{2}{c}{\textbf{Average ($\mathbf{10^{-3}}$ s)}} \\
\cmidrule(lr){2-3} \cmidrule(lr){4-5}
 {} & $k=50$ & $k=100$ & $k=50$ & $k=100$ \\
 \midrule
 Retriever (\ttt{BM25}) & 12 & 22 & 3.3 & 6.1 \\
 Ranker (\ttt{PReGAN}) & \textbf{2} & \textbf{8} & \textbf{0.5} & \textbf{2.2} \\
 Reader & 207.0 & 401.0 & 57.3 & 111.0 \\
\bottomrule
\end{tabular}
\label{tab:time_on_nqsub}
\end{table}

As we can see, the time cost of our ranker is very small compared to the retrieval and reading steps. The reader is the most expensive component. The more we submit passages to the reader, the more time it takes to find the answers. Therefore, it is advantageous to submit only a few passages to the reader without alternating the quality of final answers, which our ranker is able to do (see Figure~\ref{topk:a}).

\subsubsection{{Effects of Discriminators}}
We used two discriminators together for different criteria. We show in Table~\ref{tab:nodis} the performance of the ranker  without the answer discriminator component (\ttt{w/o AD}).
We can see that the HITS measures generally decrease when the answer discriminator is removed.
This confirms the utility of this additional discriminator in the GAN framework. 
\begin{table}[t]
    \small
    \centering
    \caption{Performance of ranker without answer discriminator (AD) on Quasar-T. 
    }
    \begin{tabular}{lccccc}
    \toprule
         & Hits@1 & Hits@3 & Hits@5 & Hits@20 & Hits@50 \\
    \midrule
         \ttt{PReGAN} & \textbf{35.2} & \textbf{52.0} & \textbf{59.5} & \textbf{72.3}  & \textbf{74.8}  \\
         \quad \ttt{w/o AD} & 33.0 & 50.1 & 58.1 & 72.0 & 74.8 \\
    \bottomrule
    \end{tabular}
    \label{tab:nodis}
\end{table}

\subsubsection{Case Study}
Let us analyze the example we gave at the beginning of the paper (Figure~\ref{fig:example}) to see the impact of our ranker on some concrete example. The example is from the  Quasar-T dataset. We use the \ttt{DPR} retriever and our ranker (on top of \ttt{BM25} results) to score the candidate passages. In this example, the passage $\mathrm{P_3}$ can be easily ranked low by both approaches due to its low relevance to the question.
The passage $\mathrm{P_4}$ is the best passage that contains the right answer to the question, and the passage provides explicit support for the answer. $\mathrm{P_2}$ is topically relevant and also contains the answer, but the passage does not provide support for the answer.
Between the passages $\mathrm{P_2}$ and $\mathrm{P_4}$, \ttt{DPR} prefers $\mathrm{P_2}$ because it contains more question-related words, \ie, more topically relevant. Topical relevance is indeed the major criterion used in \ttt{DPR}'s retriever, as can be reflected in the ranking of the passages. On the other hand, our ranker can successfully rank the best passage $\mathrm{P_4}$ on top, due to its integration of some reading capability. When a reader goes through the two passages, it is expected that the reader will find the answer in $\mathrm{P_4}$ with a higher probability than in $\mathrm{P_2}$.
This example makes the expected impact of our ranker more concrete. The effect is exactly what we desire.

One could argue that the ranking difference of the desired passage is not so important as long as the passage is submitted to the reader, and we could simply rely on the capability of the reader to find the right answer. This is not true because: 1) ranking a good passage at a lower rank will make it more likely to be cut off (not submitted to the reader), especially when the reader can only read a small number of passages; 2) the final answer from the reader also takes into account the ranking score -- the more the ranker is confident about the passage, the more the reader will have confidence about the answer in the passage.  So the correct ranking score will directly impact the final answer of the reader.

\section{Conclusion}
In this paper, we addressed the problem of passage (re-)ranking for open-domain QA. The goal is to create a better and reduced ranking list for the reader. Previous attempts on passage ranking have focused on improving the relevance of the passages~\cite{karpukhin2020dense}, using explicit interaction between the ranker and the reader~\cite{wang2018r} or incorporating some reading capability into the ranker~\cite{lin2018denoising}. However, given the noisy nature of the training data available, these approaches are hindered by the noise. In this paper, we proposed a model based on a customized GAN framework to support answer-oriented passage ranking. We utilized two discriminators -- ranking discriminator and answer discriminator, to guide the reading-based generator to assign high probabilities to passages that are relevant and contain the answer. Experiments on five datasets demonstrated the effectiveness of our proposed method. 

The work can be further improved on several aspects. First, due to the computation resources, we were unable to do experiments on large datasets (such as the full set of Natural Questions). We are looking for more powerful computation servers to run larger experiments. Second, our design choice has been inspired by the previous studies (namely, \ttt{DSQA}), which uses a simpler architecture (biLSTM) than the one commonly used now (based on \ttt{BERT}). More recent approaches are typically based on BERT. We expect that similar improvements will obtain from the idea of incorporating both relevance and answerability into a ranker, even though another scoring network is used. This will be tested in the future. 

\appendix
\section{Appendix}
\subsection{Derivation of the Reward Function}\label{sec:reward}
The detailed derivation process of the reward functions and regularizers are as follows:
\begin{equation}\label{eq:policy1}
    \begin{split}
        \nabla_\theta~& \mathcal{L}^G (q_n) \\
        =~& \nabla_\theta \mathbb{E}_{d\sim p_\theta}\left[\log(1+\exp(f_\phi(d,q_n)))\right] + \\
        & \lambda_1 \nabla_\theta \mathbb{E}_{d\sim p_\theta}\left[\log(1+\exp(f_\xi(d,q_n)))\right] - \\
        & \lambda_2 \nabla_\theta \mathbb{E}_{d\sim p_{true}}[\log p_\theta(d|q_n, a)]\\
        =~& \sum_{i=1}^M\nabla_\theta p_\theta (d_i|q_n, a) \log(1+\exp(f_\phi(d_i,q_n))) + \\
        & \lambda_1 \sum_{i=1}^M\nabla_\theta p_\theta (d_i|q_n,a) \log(1+\exp(f_\xi(d_i,q_n))) - \\
        & \lambda_2 \sum_{i=1}^M\nabla_\theta p_{true}(d_i|q_n,a)\log p_\theta(d_i|q_n,a) \\
        =~& \sum_{i=1}^M p_\theta(d_i|q_n,a)\nabla_\theta\log p_\theta(d_i|q_n,a) \log(1+\exp(f_\phi(d_i,q_n))) +\\
        & \lambda_1 \sum_{i=1}^M p_\theta(d_i|q_n,a)\nabla_\theta\log p_\theta(d_i|q_n,a) \log(1+\exp(f_\xi(d_i,q_n))) -\\
        & \lambda_2 \sum_{i=1}^M\nabla_\theta p_{true}(d_i|q_n,a)\log p_\theta(d_i|q_n,a) \\
        =~& \mathbb{E}_{d\sim p_\theta} \Big[\nabla_\theta \log p_\theta(d|q_n,a) \Big( \log(1+\exp(f_\phi(d,q_n))) + \\ ~~& \lambda_1~\log(1+\exp(f_\xi(d,q_n))) \Big) \Big] - \lambda_2 \mathbb{E}_{d\sim p_{true}}\Big[\nabla_\theta \log p_\theta(d|q_n,a)\Big] \\
        =~& \frac{1}{K} \sum_{k=1}^K\nabla_\theta \log p_\theta(d_k|q_n,a)\Big(\log(1+\exp(f_\phi(d_k,q_n)))+\\ 
        &\lambda_1~\log(1+\exp(f_\xi(d_k,q_n))) \Big) -\lambda_2 \sum_{m=1}^M \frac{p_{true}(d_m|q_n,a)}{ p_\theta(d_m|q_n, a)}
    \end{split}
\end{equation}
where the term $\log(1+\exp(f_\phi(q_n,a))) + \lambda_1 \log(1+\exp(f_\xi(q_n,a)))$ acts as the balanced rewards from both of the rank discriminator $D_\phi^r$ and the answer discriminator $D_\xi^a$.

To reduce the variance during the reinforcement learning process, the reward function $f_{rw}$ is usually modified in policy gradient in practice as below:
\begin{equation}
\begin{split}
    f_{rw} = & \log(1+\exp(f_\phi(q_n,a))) \\
    &+ \lambda_1 \log(1+\exp(f_\xi(q_n,a))) \\
    &- \mathbb{E}_{d\sim p_\theta}\Bigg[\log(1+\exp(f_\phi(q_n,a))) \\
    &+ \lambda_1 \log(1+\exp(f_\xi(q_n,a)))\Bigg].
\end{split}
\end{equation}

\balance
\bibliographystyle{ACM-Reference-Format}
\bibliography{PReGAN}
\end{document}